\pdfoutput=1

\documentclass[11pt]{article}

\usepackage[final]{ACL2023}

\usepackage{times}
\usepackage{latexsym}

\usepackage[T1]{fontenc}

\usepackage[utf8]{inputenc}

\usepackage{microtype}

\usepackage{inconsolata}
\usepackage{enumitem}
\usepackage{amsmath,amssymb}
\usepackage{algorithm}    
\usepackage{algorithmic}  
\usepackage{booktabs}
\usepackage{graphicx}
\usepackage{listings}
\usepackage{xcolor}
\usepackage{comment}
\usepackage[capitalize, nameinlink]{cleveref}

\definecolor{codegreen}{rgb}{0,0.6,0}
\definecolor{codegray}{rgb}{0.5,0.5,0.5}
\definecolor{codepurple}{rgb}{0.58,0,0.82}
\definecolor{backcolour}{rgb}{0.95,0.95,0.92}

\lstdefinestyle{mystyle}{
    backgroundcolor=\color{backcolour},
    commentstyle=\color{codegreen},
    keywordstyle=\color{magenta},
    numberstyle=\tiny\color{codegray},
    stringstyle=\color{codepurple},
    basicstyle=\ttfamily\footnotesize,
    breakatwhitespace=false,
    breaklines=true,
    captionpos=b,
    keepspaces=true,
    numbers=left,
    numbersep=5pt,
    showspaces=false,
    showstringspaces=false,
    showtabs=false,
    tabsize=2
}

\newcommand{\mee}[1]{\textcolor{red}{AA: #1}}

\title{SymCode: A Neurosymbolic Approach to Mathematical Reasoning via Verifiable Code Generation}

\author{
  Sina Bagheri Nezhad$^{1,2}$, 
  Yao Li$^1$, 
  Ameeta Agrawal$^1$
  \\ \vspace{1mm} 
  $^1$Portland State University, Portland, USA \\
  $^2$ElastixAI, Seattle, USA
  \\ \vspace{1mm} 
  \texttt{\{sina5, liyao, ameeta\}@pdx.edu}
}

\begin{document}
\maketitle
\begin{abstract}
Large Language Models (LLMs) often struggle with complex mathematical reasoning, where prose-based generation leads to unverified and arithmetically unsound solutions. Current prompting strategies like Chain of Thought still operate within this unreliable medium, lacking a mechanism for deterministic verification. To address these limitations, we introduce SymCode, a neurosymbolic framework that reframes mathematical problem-solving as a task of verifiable code generation using the SymPy library. We evaluate SymCode on challenging benchmarks, including MATH-500 and OlympiadBench, demonstrating significant accuracy improvements of up to 13.6 percentage points over baselines. Our analysis shows that SymCode is not only more token-efficient but also fundamentally shifts model failures from opaque logical fallacies towards transparent, programmatic errors. By grounding LLM reasoning in a deterministic symbolic engine, SymCode represents a key step towards more accurate and trustworthy AI in formal domains.
\end{abstract}

\section{Introduction}


Large Language Models (LLMs) have demonstrated remarkable capabilities in natural language, yet their proficiency in domains requiring rigorous, multi-step formal reasoning, such as advanced mathematics, remains a significant challenge \citep{wei2022chain, ahn-etal-2024-large}. When prompted to solve complex math problems, LLMs that reason in prose often generate solutions that are unreliable, containing subtle arithmetic errors, logical fallacies, or hallucinated intermediate steps. Furthermore, these natural language rationales are often opaque; their convoluted structure can obscure the reasoning path, making it difficult for even a domain expert to verify their correctness. This lack of a clear, deterministic failure signal also makes it challenging to create an automated feedback loop to iteratively refine an incorrect answer.

Current approaches to mathematical reasoning broadly fall into two categories: inference-time prompting and model fine-tuning. Inference-time methods like Chain of Thought (CoT)~\citep{wei2022chain} and Tree of Thoughts (ToT)~\citep{yao2023tree} have improved performance by encouraging models to articulate their reasoning. These methods, while accessible, inherit the weaknesses of natural languages. Training-based methods, on the other hand, can improve a model's intrinsic capabilities but require significant computational resources and large, high-quality datasets, and may not generalize well to novel problems.

To overcome these critical shortcomings, we introduce \textbf{SymCode}, a neurosymbolic framework that reframes mathematical problem-solving for any class of problems that can be formalized programmatically. Instead of prompting an LLM to describe its reasoning in prose, SymCode instructs it to construct a verifiable, executable \textit{Python script} where the code serves as the reasoning trace. Unlike prior work like Program-Aided Language Models (PAL) \citep{gao2023pal}, which uses code as an external calculator for intermediate steps, SymCode treats the \textit{entire program} as the final, self-contained reasoning artifact. This elevates the LLM's role from a simple calculator to an expert translator, converting a natural language problem into a formal, verifiable script.

\begin{figure*}[!t]
    \centering
    \includegraphics[width=\textwidth, trim=0 18 0 0, clip]{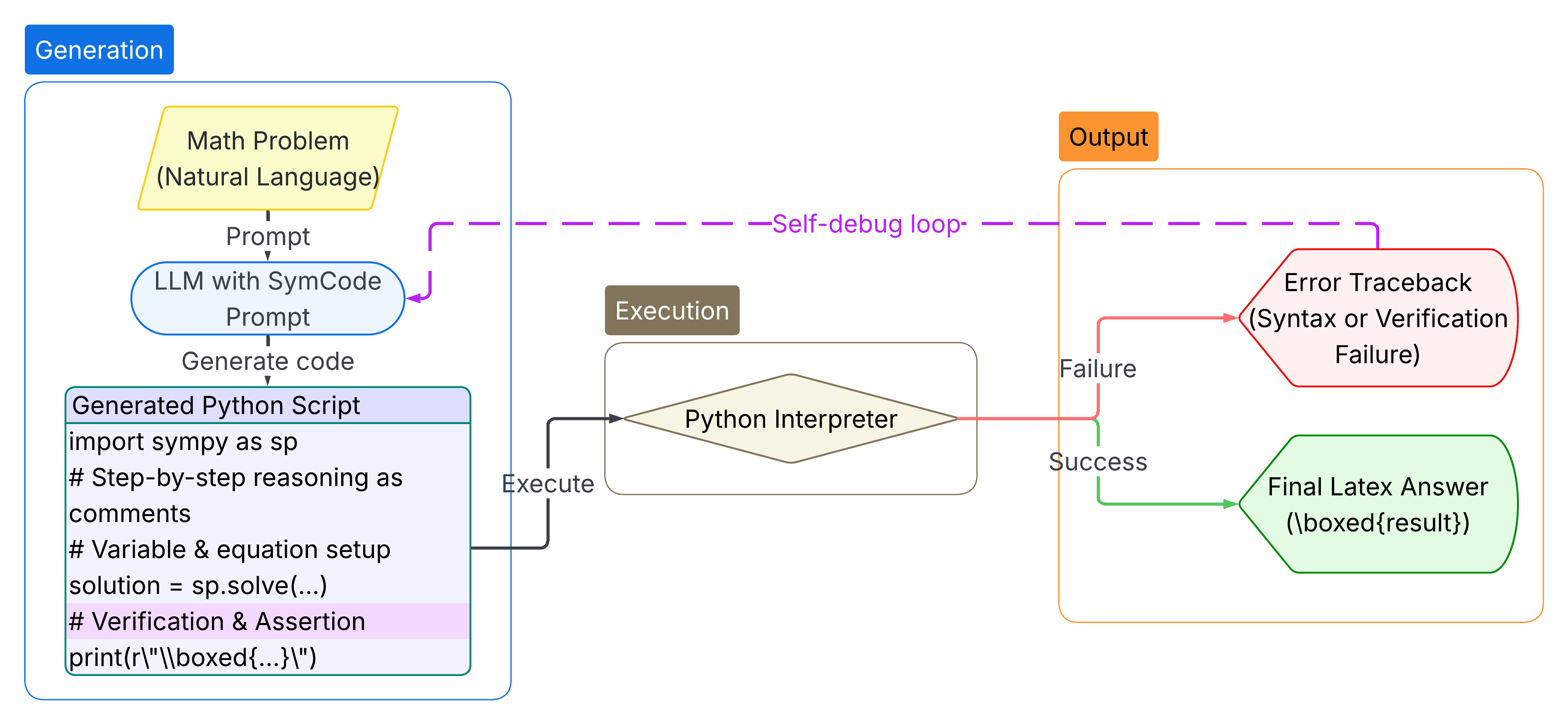}
    \caption{Overview of the SymCode framework. A natural language problem is translated into Python code by the LLM, executed, and iteratively refined through error feedback until successful execution or a retry limit is reached.} 
    \label{fig:framework_diagram}
\end{figure*}

The SymCode framework orchestrates the strengths of three components to address the challenges of prose-based reasoning. As illustrated in Figure~\ref{fig:framework_diagram}, the process begins with instructing an LLM to interpret the problem and generate a Python script that leverages \texttt{SymPy}, a computer algebra system~(CAS) that manipulates mathematical expressions in their exact symbolic form, thereby eliminating arithmetic errors. Next, the generated script is executed in a sandboxed Python interpreter, which provides a deterministic pass/fail signal for programmatic verification. Finally, a self-debugging loop feeds any execution error back to the LLM, enabling it to iteratively correct its own code. This entire process creates a transparent, auditable, and self-correcting reasoning trace that is both human-readable and machine-executable.

Our work is guided by two main research questions:
\begin{enumerate}
    \item To what extent does reframing mathematical reasoning as verifiable code generation improve a system's accuracy on complex mathematical problems compared to established prose-based prompting techniques?
    \item Beyond accuracy, how does this neurosymbolic approach alter the characteristics of the LLM's reasoning process, specifically concerning token efficiency and the fundamental nature of its failure modes?
\end{enumerate}

Our main contributions are as follows:
\begin{itemize}
    \item We introduce and formalize \textbf{SymCode}, a prompt-based framework for advanced mathematical reasoning that transforms an LLM into a neurosymbolic reasoner generating self-contained, verifiable Python scripts.
    \item Framing mathematical reasoning as code generation enables us to apply an iterative self-debugging mechanism where LLM uses deterministic interpreter feedback to correct its own errors, a robust verification process not available to prose-based reasoners.
    \item Through extensive experiments on three challenging mathematical benchmarks—MATH-500, OlympiadBench, and AIME—we show that SymCode improves accuracy by up to 13.6 percentage points (and up to 16.8 with SymCode+) over traditional prompting baselines, with the performance gap increasing as problem difficulty rises.
    \item We provide a detailed analysis showing that reasoning-as-code is substantially more token-efficient than prose-based methods.
\end{itemize}


\section{Related Work}
\begin{table*}[h!]
\setlength{\tabcolsep}{5pt}
\centering
\small
\begin{tabular}{lccccc}
\toprule
\textbf{Method} & \textbf{Reasoning Modality} & \textbf{External Tools} & \textbf{Self-Correction} & \textbf{Verification} \\
\midrule
\multicolumn{5}{l}{\textit{\textbf{Training-Based Methods (Model-Tuning)}}} \\
\midrule
LeDex \citeyear{10.5555/3737916.3739036} & Code & Interpreter Feedback & Yes (Learned) & Execution-based \\
rStar-Math \citeyear{guan2025rstarmathsmallllmsmaster} & Hybrid (Code+) & MCTS+Code & Yes (Self-Evolution) & Process Reward Model \\
\midrule
\multicolumn{5}{l}{\textit{\textbf{Training-Free Methods (Inference-Time)}}} \\
\midrule
CoT \citeyear{wei2022chain} & Natural Language & None & No & No \\
ToT \citeyear{yao2023tree} & Natural Language & None & Branching & None (NL-based vote) \\
PAL \citeyear{gao2023pal} & Code & Python Interpreter & No & Result Only \\
MATHSENSEI \citeyear{das2024mathsenseitoolaugmentedlargelanguage} & Hybrid & Search/Prog/Solver & No (has code refiner) & Tool-based \\
NSAR \citeyear{bagheri2025enhancing} & Hybrid & Python/Symbolic & No & Result Only \\
\textbf{SymCode+ (Ours)} & \textbf{Code} & \textbf{SymPy (CAS)} & \textbf{Yes (Self-Debug Loop)} & \textbf{Constraints and Result} \\

\bottomrule
\end{tabular}
\caption{Comparison of SymCode+ with key LLM-based mathematical reasoning methods, grouped into training-free inference-time approaches and training-based fine-tuning approaches. SymCode is a training-free method focused on verifiable \texttt{SymPy} code generation with a self-debugging loop. Abbrev.: CAS = Computer Algebra System, MCTS = Monte Carlo Tree Search.}
\label{tab:related-work-comparison} 
\end{table*}

The use of code for mathematical reasoning is well-established, from the computer-assisted proof of the four-color theorem \citep{appel1989every} to modern formal proof assistants like Lean \citep{moura2021lean}. Historically, however, these powerful symbolic systems required expert human effort to manually translate natural language problems into formal specifications. The recent challenge of automating this translation and equipping LLMs with robust mathematical abilities has spurred a variety of research directions, which can be broadly categorized into three main areas: training-based methods that modify model weights, training-free strategies that operate at inference-time, and methods that augment LLMs with external tools.

A significant line of research focuses on enhancing reasoning by fine-tuning models on specialized data or with reinforcement learning. For instance, \texttt{rStar-Math} employs Monte Carlo Tree Search (MCTS) guided by a process reward model to create a ``deep thinking'' process, fine-tuning smaller models to achieve strong performance \citep{guan2025rstarmathsmallllmsmaster}. These training-based methods represent a powerful but distinct paradigm focused on improving the model's internal capabilities. 


Another category of methods aims to improve reasoning without altering the model's parameters, focusing instead on structuring the generation process at inference-time. Early efforts in this area focused on eliciting more structured thought processes. The seminal Chain of Thought (CoT) prompting method demonstrated that instructing a model to ``think step-by-step'' significantly improves performance \citep{wei2022chain}. This concept was further generalized by approaches like Tree of Thoughts (ToT), which allows models to explore multiple reasoning paths concurrently \citep{yao2023tree}, and decomposition techniques that break complex problems into simpler sub-problems \cite{khot2023decomposedpromptingmodularapproach}. More recently, this paradigm has shifted towards scaling test-time compute, a strategy popularized by OpenAI's \texttt{o1} model \cite{openai2024openaio1card}. This has led to methods like \textit{budget forcing}, where a model's thinking process is deliberately extended to encourage deeper exploration \citep{muennighoff2025s1simpletesttimescaling}. While effective, these methods still largely operate within the domain of natural language, making their reasoning chains prone to arithmetic errors and logical inconsistencies without a formal method for verification \citep{ahn-etal-2024-large}.

To overcome the unreliability of prose-based computation, a third line of work has focused on augmenting LLMs with external tools, particularly code interpreters. Program-Aided Language Models (PAL) \citep{gao2023pal} pioneered this approach by prompting an LLM to generate an executable program, offloading computation to a reliable interpreter. This has been extended with models like MATHSENSEI, which integrates multiple tools including web search and symbolic solvers \citep{das2024mathsenseitoolaugmentedlargelanguage}. Our work represents a fundamental shift from this paradigm. Rather than using code for mere computation, we use it to change the reasoning modality itself. \texttt{SymCode} instructs the LLM to translate a problem into a formal, \textit{symbolic representation}, which is then manipulated by a Computer Algebra System (CAS). This elevates the task from executing a sequence of arithmetic steps to solving a system of symbolic equations. In essence, PAL uses code for calculation, whereas \texttt{SymCode} uses code for formal mathematical reasoning. This directly operationalizes a true neurosymbolic approach by bridging neural language interpretation with the rigorous logic of symbolic systems \citep{kautz2022third,fang2024large,bagheri2025enhancing}.

The use of code also enables robust self-correction mechanisms. Several methods train models to debug their own code, either through fine-tuning on datasets of errors and corrections \citep{10.5555/3737916.3739036} or by using reinforcement learning to refine outputs based on execution feedback \citep{kumar2024traininglanguagemodelsselfcorrect}. While some studies suggest that Reinforcement Learning with Verifiable Rewards (RLVR) \cite{lambert2025tulu} primarily amplifies existing capabilities rather than creating new ones \citep{yue2025doesreinforcementlearningreally}, the iterative self-debugging loop in \texttt{SymCode} provides a concrete, inference-time mechanism for refinement. 

Our approach also differs from general-purpose code generation, where models produce applications from specifications (e.g., text-to-SQL) \citep{zan-etal-2023-large}. In those tasks, the code is the final product. In contrast, we employ code as the \textit{reasoning modality itself}—a transparent, intermediate representation of logic. 

To highlight the distinctions and contributions of our approach, Table~\ref{tab:related-work-comparison} provides a comparative overview of key related methods in mathematical reasoning.

\section{The SymCode Framework}
The SymCode framework adapts an LLM from a probabilistic text generator into a structured, neurosymbolic reasoner. While prior work has used code as an external computational tool, the fundamental principle of SymCode is to treat the entire reasoning process as the act of generating a  verifiable program where the code \textit{is} the reasoning trace. The motivation for this shift is to overcome the inherent limitations of prose-based reasoning, which is often ambiguous, prone to subtle logical and arithmetic errors, and lacks a mechanism for automated verification. Instead of asking the LLM to explain its thinking, we instruct it to write a program that \textit{enacts} that thinking. This methodology leverages the respective strengths of neural and symbolic systems: the LLM excels at interpreting the nuances and context of the problem statement, while the Python interpreter, coupled with the \texttt{SymPy} library \cite{10.7717/peerj-cs.103}, provides rigor for the formal mathematical manipulations. First, we use a specialized prompt to guide the LLM in structuring its reasoning as a self-contained Python script that leverages the \texttt{SymPy} library for deterministic computation~(\cref{subsec:prompt}). Second, to enhance accuracy, we introduce an iterative self-debugging loop that enables the model to correct its own programmatic errors based on interpreter feedback~(\cref{subsec:symcode-plus}). To provide a concrete illustration of the framework in action, from problem statement to the final generated script, see the full example in Appendix \ref{sec:full_example}.



\subsection{SymCode}\label{subsec:prompt}
The complete SymCode prompt template is shown below.

\begin{lstlisting}[caption={The SymCode Prompt Template.}, label={lst:prompt}, numbers=none]
You are an expert mathematical reasoner. Your output must be ONLY a single Python code block fenced as ```python ... ``` with no prose before or after.
Inside that single Python script:
1. Import SymPy with `import sympy as sp`
2. Add explicit step-by-step reasoning as comments throughout your code
3. Document the problem setup:
   - Clearly identify variables, constraints, and goals in comments
   - Define symbols with appropriate assumptions (e.g., sp.symbols('x', positive=True, integer=True))
4. Include intermediate reasoning steps:
   - Each step should have a comment explaining the mathematical reasoning
   - Use meaningful variable names that reflect their purpose
   - Show the algebraic manipulations clearly
5. For verification:
   - Substitute solutions back into original equations
   - Check domain constraints (e.g., integer solutions, positive values)
   - Filter invalid solutions
6. Print ONLY the final answer in LaTeX boxed form:
   print(r"\boxed{}".format(final_answer))
# PROBLEM
{problem_text}
# END PROBLEM
\end{lstlisting}

\begin{table*}[!t]
\centering
\resizebox{\textwidth}{!}{%
\begin{tabular}{@{}llcccc@{}}
\toprule
\textbf{Dataset} & \textbf{Problem Statement} & \textbf{Ground Truth} & \textbf{SymCode} & \textbf{CoT} & \textbf{ToT} \\ \midrule
\textbf{MATH-500} & \begin{tabular}[c]{@{}l@{}}How many distinct values can be obtained \\ from the expression $2\cdot 3\cdot 4 \cdot 5 + 1$ by \\ inserting any number of parentheses?\end{tabular} & 4 & 4 & 1 & 2 \\ \midrule
\textbf{OlympiadBench} & \begin{tabular}[c]{@{}l@{}}In $\triangle ABC$, $AB=4, BC=6, AC=8$. \\ Squares $ABQR$ and $BCST$ are drawn external \\ to the triangle. Compute the length of $QT$.\end{tabular} & $2\sqrt{10}$ & $2\sqrt{10}$ & $2\sqrt{2}\sqrt{10+3\sqrt{15}}$ & 12 \\ \midrule
\textbf{AIME} & \begin{tabular}[c]{@{}l@{}}Let $\triangle ABC$ have circumcenter $O$ and incenter \\ $I$ with $\overline{IA}\perp\overline{OI}$, circumradius $13$, and \\ inradius $6$. Find $AB\cdot AC$.\end{tabular} & 468 & 468 & 156 & 338 \\ \bottomrule
\end{tabular}%
}
\caption{Sample problems from the evaluation datasets, with outputs from SymCode and prose-based baselines. SymCode correctly solves all three, while the baselines produce incorrect answers due to logical or arithmetic errors.}
\label{tab:sample-outputs}
\end{table*}

Each component of this prompt serves a distinct purpose in structuring the model's output.

\paragraph{Symbolic Formulation.} The explicit instruction to use \texttt{import sympy as sp} is critical. \texttt{SymPy} is a Python library for symbolic mathematics, acting as a Computer Algebra System (CAS). Unlike standard numerical libraries that work with approximate floating-point numbers, \texttt{SymPy} manipulates mathematical expressions in their exact, symbolic form (e.g., representing $\sqrt{2}$ precisely rather than as 1.414...). This brings two key advantages: first, it prevents the accumulation of rounding errors that can invalidate multi-step calculations. Second, it enables true algebraic reasoning by allowing the script to programmatically solve equations, simplify expressions, and apply mathematical rules with perfect fidelity. This moves the task from error-prone prose-based calculation to exact, verifiable computation.

\paragraph{Interpretability.} By requiring `step-by-step reasoning as comments', we retain the ``show your work'' benefit of Chain of Thought while grounding it in a formal, code-based structure. This makes the model's logic transparent and auditable for human experts.

\paragraph{Problem Scaffolding.} The requirement to define symbols with appropriate assumptions (e.g., `sp.symbols(`x', positive=True, integer=True)') forces the model to formalize the problem's constraints upfront. This structured setup significantly reduces the solution search space and helps prevent the generation of invalid solutions later in the process.

\paragraph{Verification and Filtering.} This is a cornerstone of the framework's reliability. The prompt requires the model to insert \texttt{assert} statements into its generated code. These statements check key conditions at runtime, such as whether a solution satisfies the original problem constraints or adheres to domain requirements (e.g., ensuring a length variable is positive). If an assertion fails, it raises an error that halts execution. This provides a deterministic failure signal that effectively filters out incorrect solution paths and can trigger the self-debugging loop of SymCode+ (described next), enabling a crucial self-correction step. If the script runs to completion without any exceptions, the final answer is reported by capturing the script's output, which the prompt requires to be printed in a \texttt{\textbackslash boxed\{\}} LaTeX format. The effectiveness of this verification, however, is contingent on the quality of the assertions generated by the LLM; an absence of failure does not guarantee correctness if the assertions are weak or located in an unexecuted code path.

\subsection{SymCode+: Self-Debugging Loops}\label{subsec:symcode-plus}
To further enhance the accuracy of the framework, we extend the core prompt with an agentic wrapper called \textbf{SymCode+}. This extension introduces an iterative self-debugging loop that allows the LLM to correct its own programmatic mistakes.

If the initial script fails during execution, the loop is activated. A failure can be a programmatic \textit{exception} (e.g., \texttt{SyntaxError}, \texttt{TypeError}) or a \textit{verification failure} where an internal \texttt{AssertionError} is raised. The captured error message and traceback are then appended to the prompt history, and the LLM is instructed to ``debug the following code based on the provided error message.'' This cycle of execution, failure, and correction repeats until the script runs successfully or a preset iteration limit (e.g., 2--3 attempts) is reached.


\section{Experimental Setup}

We consider challenging mathematical datasets and a diverse set of LLMs in our evaluation.

\subsection{Datasets}

We evaluate SymCode on three widely used, challenging benchmarks that require multi-step mathematical reasoning, spanning difficulty from high school competitions to Olympiad-level problems: 
\noindent \textbf{(1) MATH-500}, a 500-problem subset of the MATH dataset, comprising challenging problems from high school mathematics competitions, covering topics like algebra, geometry, number theory, and precalculus \cite{lightman2023letsverifystepstep}; \\
\noindent \textbf{(2) OlympiadBench}, 674 text-only English math problems from national and international olympiads requiring creative and formal reasoning \cite{he-etal-2024-olympiadbench}; and\\
\noindent \textbf{(3) American Invitational Mathematics Examination (AIME) 2024 \& 2025}, 60 (30 from each year) short-answer problems from recent competitions that bridge high school and olympiad difficulty \cite{maa2024aime}. \Cref{tab:sample-outputs} shows a sample from each dataset.


\subsection{Models and Baselines}
We evaluate SymCode across several state-of-the-art language models such as Llama 3.2 (90B)\cite{grattafiori2024llama3herdmodels}, GPT-5-nano \cite{openai2025gpt5}, and GPT-OSS (20B) \cite{openai2025gptoss120bgptoss20bmodel}  (reasoning level “high”), chosen to span a range of coding fluency, from strong code generators (GPT-5-nano, GPT-OSS) to a generalist model less optimized for coding (Llama 3.2). Focusing on small and medium models rather than frontier-scale systems (e.g., GPT-5, Grok 4) allows us to investigate more efficient, accessible approaches to improving reasoning performance.

The performance of SymCode is contextualized against a set of strong, widely-used baseline prompting strategies, including:

\begin{itemize}[leftmargin=*]
    \item \textbf{Chain of Thought (CoT):} A standard baseline where the model is prompted to ``think step-by-step'' to generate a prose-based rationale before giving the final answer \cite{wei2022chain}.
    \item \textbf{Tree of Thoughts (ToT):} An advanced baseline where the model explores multiple reasoning paths, evaluating and pruning them to find the most promising solution \cite{yao2023tree}.
    \item \textbf{Decomposition:} A baseline where the model is instructed to break the problem into smaller, simpler sub-problems and solve them sequentially \cite{khot2023decomposedpromptingmodularapproach}.
\end{itemize}

Because SymCode is a training-free framework that modifies reasoning only at inference time, training-based methods are not directly comparable baselines, as fair comparison would require adapting our approach to their specialized models. We also exclude code-generation methods like PAL \cite{gao2023pal}, whose main purpose is delegating numerical computation to an interpreter. Furthermore, our initial exploratory tests confirmed that PAL is of limited utility for the complex problems in our benchmarks, as it is primarily designed for numerical outputs and struggles significantly with problems requiring a final symbolic expression as the answer.

\begin{table*}[!t]
\centering
\small
\begin{tabular}{llccc}
\toprule
\textbf{Model} & \textbf{Method} & \textbf{MATH-500} & \textbf{OlympiadBench} & \textbf{AIME (24-25)} \\
\midrule
\textbf{Llama 3.2 (90B)} & CoT & 61.2 & 34.4 & 20.0 \\
                         
                         & ToT & 63.8 & \textbf{36.8} & 23.3 \\
                         & Decomposition & \underline{64.4} & \underline{36.0} & 21.7 \\
                         & SymCode (ours) & \underline{64.4} & 31.2 & \underline{25.0} \\
                         & SymCode+ (ours) & \textbf{68.8} & \textbf{36.8} & \textbf{31.7} \\
\midrule
\textbf{GPT-5-nano} & CoT & \textbf{93.4} & 63.2 & 51.6 \\
                    
                    & ToT & 88.2 & 68.0 & 51.7 \\
                    & Decomposition & 91.2 & 64.0 & 48.3 \\
                    & SymCode (ours) & 90.8 & \underline{76.8} & \underline{61.7} \\
                    & SymCode+  (ours) & \underline{91.4} & \textbf{80.0} & \textbf{65.0} \\
\midrule
\textbf{GPT-OSS (20B)} & CoT & 88.4 & 22.4 & 11.6 \\
                       
                       & ToT & 87.0 & 24.8 & 10.0 \\
                       & Decomposition & 86.2 & 23.2 & 11.6 \\
                       & SymCode (ours) & \underline{90.4} & \underline{35.2} & \underline{18.3} \\
                       & SymCode+ (ours) & \textbf{93.2} & \textbf{38.4} & \textbf{21.6} \\
\bottomrule
\end{tabular}
\caption{Accuracy (\%) on Mathematical Reasoning Benchmarks. Best score shown in {\bf bold} whereas second best score is \underline{underlined}.}
\label{table:accuracy}
\end{table*}

\subsection{Evaluation Metrics}
We assess SymCode using four metrics: (1) {\bf Accuracy}: a solution is correct only if the final numerical or symbolic answer inside the `boxed\{\}` output exactly matches the ground-truth solution, no partial credit is awarded; (2) {\bf Token Efficiency}: the average tokens generated per problem, reflecting the conciseness of code-based reasoning; (3) {\bf Qualitative Error Analysis}: manual categorization of failure types, contrasting arithmetic or logical errors in baselines with misinterpretation or API errors in SymCode; and (4) {\bf Self-Debugging Activation Rate}: the percentage of problems triggering the self-correction loop, indicating the model’s initial coding fluency.

\section{Results and Analysis}
This section presents a detailed discussion of the experimental results.

\subsection{Overall Performance and the Role of Coding Proficiency}
Our experiments reveal that the SymCode framework's effectiveness depends on the base model's coding proficiency and the complexity of the reasoning task. As shown in Table~\ref{table:accuracy}, SymCode delivers substantial gains, and with its self-debugging loop, SymCode+ consistently yields additional gains on the most challenging benchmarks. Accuracy results show that SymCode's advantages are most pronounced with models that are strong coders: with GPT-5-nano, SymCode+ achieves an accuracy of 80\% on OlympiadBench and 65\% on AIME, representing a remarkable absolute improvement of nearly 12 and 13 percentage points, respectively, over the best-performing prose-based baseline (ToT), demonstrating that  shifting the reasoning modality from prose to a formal symbolic language unlocks new capabilities.

With Llama 3.2 (90B), a less code-optimized model, standard SymCode underperforms ToT on OlympiadBench, but SymCode+ closes this gap, achieving 21.6\% accuracy on AIME. This shows that iterative self-correction is crucial for models with weaker coding skills, allowing them to overcome initial syntax or logic errors.

A closer look at the model-specific results reveals interesting trade-offs between coding proficiency and abstract reasoning. As expected, {GPT-5-nano} stands out as the top-performing model, demonstrating superior capabilities across the board, particularly on the most difficult OlympiadBench and AIME datasets. More intriguing is the comparison between {GPT-OSS} and {Llama 3.2 (90B)}. While GPT-OSS shows stronger performance on the MATH-500 benchmark (93.2\% vs. 68.8\%), Llama 3.2 surprisingly surpasses it on the more challenging AIME problems (31.7\% vs. 21.6\%). A potential explanation for this reversal lies in the nature of the tasks. MATH-500 problems, while complex, are often more standard in structure, allowing GPT-OSS to better leverage its strong coding abilities to translate familiar problem types into reliable scripts. In contrast, AIME problems frequently require more novel or abstract initial insights before a solution can be formalized. Llama 3.2, while a less fluent coder, appears more adept at forming the correct initial conceptual plan for these non-standard problems, even if its first attempt at implementation contains errors that the debugging loop must then correct.


\subsection{Impact of Problem Difficulty}

\begin{figure}[!ht]
    \centering
    \includegraphics[width=0.48\textwidth]{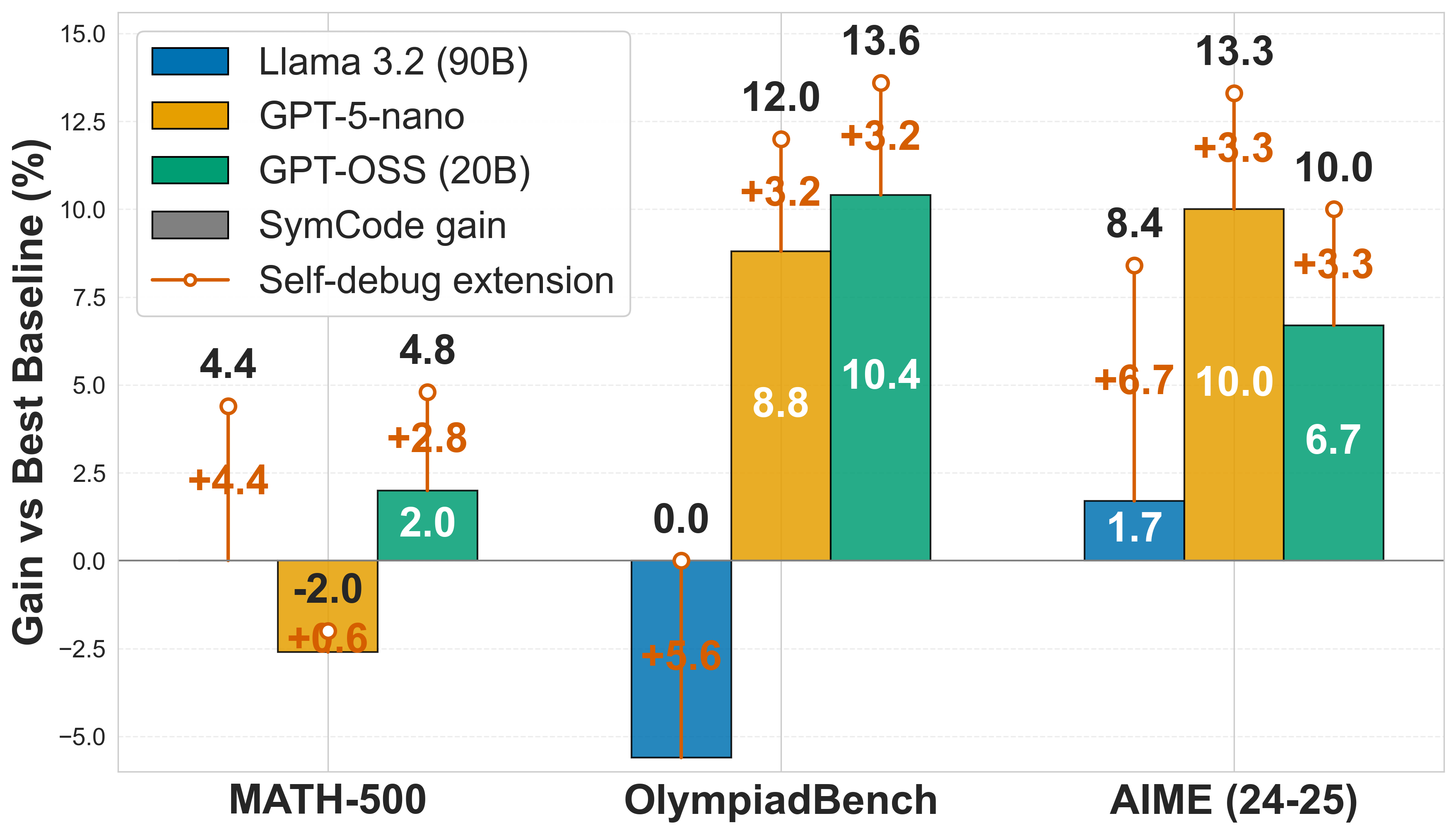}
    \caption{Performance gain of SymCode and SymCode+ over the best prose-based baseline. The advantage of the SymCode framework is most significant on the most difficult datasets (OlympiadBench and AIME).}
    \label{fig:performance_gain}
\end{figure}

A central hypothesis of our work is that the benefits of a verifiable, code-based reasoning process become more pronounced as problem complexity increases. The results strongly validate this claim. As visualized in Figure \ref{fig:performance_gain}, we plot the absolute accuracy improvement of SymCode+ over the strongest prose-based baseline for each model.

For all models, the most significant gains are on the AIME and OlympiadBench datasets, which feature problems requiring deep insight and long, precise reasoning chains. For GPT-5-nano, the gain escalates from a slight deficit on MATH-500 to a massive +13.3 point advantage on AIME. This trend suggests that while prose-based methods are effective for shorter problems, they are more susceptible to accumulating subtle arithmetic or logical errors over longer reasoning chains. By delegating execution to a deterministic \texttt{SymPy} interpreter, SymCode avoids these pitfalls, making it a more robust method for tackling complex, multi-step problems.

\subsection{Token Efficiency Analysis}

Consistent with our initial hypothesis, SymCode is substantially more token-efficient than its prose-based counterparts. Python code expresses complex operations concisely, whereas natural language requires verbose descriptions. On average (across all three datasets and all three models), SymCode solutions used just \textbf{699 output tokens}\footnote{See Appendix~\ref{app:token_accounting} for the token-counting protocol and model-specific token visibility. All reported token counts are rounded.}, proving significantly more concise than the baselines: Tree of Thoughts (\textbf{1770 tokens}), Decomposition (\textbf{1962 tokens}), and Chain of Thought (\textbf{2991 tokens}). This results in a token reduction of approximately \textbf{60\% to 77\%} compared to these prose-based methods, a significant advantage for inference cost and latency. 
The self-debugging loop in \texttt{SymCode+} raises the average token count to \textbf{890}. This overhead is most pronounced for Llama 3.2, whose higher self-debugging activation rate requires more token-intensive refinement. 

\subsection{Qualitative Error Analysis}
To understand the qualitative differences behind the accuracy scores, we manually categorized the errors made by GPT-5-nano on the AIME dataset for the best-performing baseline (ToT) and our SymCode  method. 


The ToT baseline's errors stem mainly from flaws in its reasoning process, with \textbf{arithmetic mistakes} (41.4\%), where the model makes simple miscalculations and \textbf{logical fallacies} (34.5\%), where a theorem is misapplied or a step in the logic is unsound. The remaining 24.1\% of ``Other'' errors primarily consist of {incomplete solutions}, where the model fails to finish the reasoning chain, or {hallucinated constraints}, where it invents details not present in the problem.

In contrast, SymCode shifts failure modes toward the structured stage of problem setup, with most errors due to \textbf{problem misinterpretation} (56.2\%) and \textbf{incorrect API usage} (31.3\%). The final 12.5\% of ``Other'' errors are mostly {runtime issues} like infinite loops or {verification failures} where an assertion check fails. This transition from opaque reasoning errors to transparent, programmatic ones points to clearer paths for improvement through better code generation and debugging. For a direct, side-by-side comparison illustrating how a baseline method and \texttt{SymCode} fail on the same problem, see Appendix \ref{sec:failure_analysis}.

A detailed breakdown of our results on OlympiadBench shows that while accuracy improved across all subfields, the gains were most significant in {combinatorics} and {geometry}. Performance in {number theory} saw moderate improvement. In contrast, {algebra} saw the smallest gains, not because the method is less effective, but because the baseline models already exhibited a higher initial performance in this area. Furthermore, we observed that the performance uplift was substantially larger for problems requiring a final {expression} as an answer compared to those requiring a single numerical value.

\subsection{Self-Debugging Loop Activation}

\begin{figure}[ht!]
    \centering
    \includegraphics[width=0.45\textwidth]{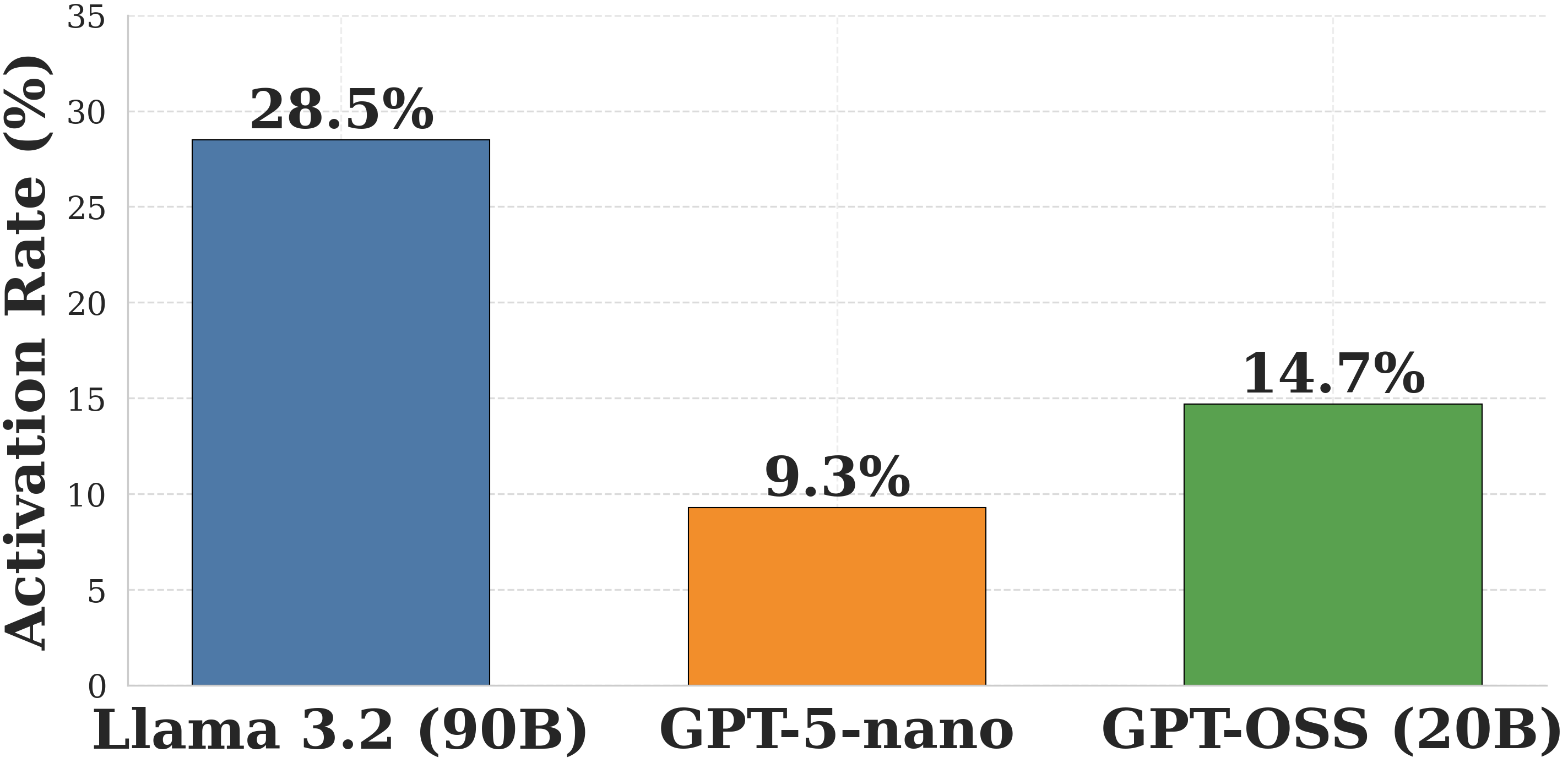}
    \caption{Activation rate of the self-debugging loop. The loop was required most often for Llama 3.2, correlating with its weaker initial coding performance.}
    \label{fig:debug_activation} 
\end{figure}

The effectiveness of the self-debugging mechanism is directly linked to the base model's coding fluency. Figure \ref{fig:debug_activation} shows the percentage of problems where the self-debugging loop was activated (i.e., the first attempt failed and a retry was initiated).

As expected, the loop was triggered most frequently for {Llama 3.2 (90B)}, with over 28\% of problems requiring at least one correction. This high activation rate correlates with its significant performance jump from standard SymCode to SymCode+ and confirms that the debugging loop is a vital component for enabling models with weaker coding skills to effectively use this framework. For the more code-proficient models, the loop was activated less often but still provided a crucial safety net for correcting errors, leading to modest but important accuracy gains.

\subsection{Ablation Analysis}

To isolate the impact of the core components of our framework, we conducted a \textit{sequential} ablation study using the best-performing model, {GPT-5-nano}, on the most challenging dataset, {AIME (24-25)}. We evaluated three progressively degraded versions of our framework, removing one key feature at each step: the iterative self-debugging loop (\textbf{No Self-Debug}), the use of assertions (\textbf{No Verification}), and the symbolic \texttt{SymPy} library in favor of standard numerical libraries (\textbf{No \texttt{SymPy} (Numeric Python)}). The results, presented in \cref{table:ablation}, confirm that each component is critical for achieving peak performance.

\begin{table}[h!]
\centering
\small
\begin{tabular}{lc}
\toprule
\textbf{Method Variant} & \textbf{AIME Accuracy (\%)} \\
\midrule
\textbf{SymCode+} & \textbf{65.0} \\
\midrule
No Self-Debug  & 61.7 \\
No Verification & 58.5 \\
No \texttt{SymPy} (Numeric Python) & 48.3 \\
\bottomrule
\end{tabular}
\caption{Ablation analysis of SymCode components on the AIME dataset using GPT-5-nano.}
\label{table:ablation} 
\end{table}

\section{Conclusion}
In this work, we introduced SymCode, a neurosymbolic framework that reframes mathematical problem-solving for large language models (LLMs) as verifiable code generation, combining neural language understanding with symbolic computation for greater precision and reliability. Experiments on challenging benchmarks like AIME and OlympiadBench show that SymCode, especially with its self-debugging loop, achieves state-of-the-art performance, with its advantage growing as problem complexity increases. By shifting reasoning from opaque text-based errors to transparent programmatic ones, SymCode enhances accuracy, efficiency, and interpretability. Looking ahead, we aim to extend this reasoning-as-code paradigm beyond mathematics to domains like physics and formal logic, and improve self-debugging through error-driven fine-tuning.

\section{Limitations}

Despite its strengths, SymCode has several limitations. Its performance depends heavily on the base LLM’s coding proficiency, with stronger code-oriented models outperforming others despite the self-debugging loop. The framework is most effective on problems that can be directly expressed symbolically, and struggles with tasks requiring abstract reasoning, such as synthetic geometry proofs, induction or contradiction, and combinatorial arguments that resist formalization in \texttt{SymPy}. Its reliability also depends on the correctness of the Python interpreter and \texttt{SymPy} library, which, while mature, are not formally verified and may propagate rare errors. Finally, SymCode relies on carefully structured prompts, making performance sensitive to prompt design and motivating future work on more robust instruction formats.

\section{Ethical Considerations and Broader Impact}

The development of advanced mathematical reasoners like SymCode has several broader implications.

\paragraph{Positive Impact} On the positive side, this technology holds significant potential for advancing scientific research and education. It could serve as a powerful assistant for scientists, engineers, and mathematicians by automating complex symbolic calculations and verifying formal proofs. In education, it could be integrated into AI tutoring systems to provide students with step-by-step, verifiable solutions to complex STEM problems, thereby enhancing learning outcomes.

\paragraph{Potential Misuse and Mitigation} Conversely, the ability to automatically solve complex mathematical problems raises concerns about academic integrity. Such a tool could be misused to cheat on assignments or standardized tests, undermining the educational process. As with any powerful AI capability, the development of SymCode must be accompanied by a broader conversation about its responsible deployment. Potential mitigation strategies include the development of specialized detectors for AI-generated code and promoting educational policies that focus on assessing the reasoning process rather than just the final answer.

\bibliography{custom}
\bibliographystyle{acl_natbib}

\appendix

\section{Full Generation Example}
\label{sec:full_example}

To provide a concrete illustration of the SymCode framework in action, this section presents a full, unedited output from the GPT-5-nano model for a representative geometry problem.

\subsection*{Problem Statement}
In triangle $A B C, A B=4, B C=6$, and $A C=8$. Squares $A B Q R$ and $B C S T$ are drawn external to and lie in the same plane as $\triangle A B C$. Compute $Q T$.

\subsection*{Generated SymCode Script}
\begin{lstlisting}[language=Python]
import sympy as sp

# Step 1: Problem understanding and geometric setup
# We place A = (0,0), B = (4,0) so AB = 4.
# Let C = (Cx, Cy) with given distances AC = 8 and BC = 6.
# We will solve for Cx, Cy from the two circle equations.

A = sp.Matrix([0, 0])
B = sp.Matrix([sp.Integer(4), 0])

Cx, Cy = sp.symbols('Cx Cy', real=True)

# Equations: |AC|^2 = 64 and |BC|^2 = 36
eq_AC = sp.Eq(Cx**2 + Cy**2, 64)
eq_BC = sp.Eq((Cx - 4)**2 + Cy**2, 36)

# Step 2: Solve for Cx, Cy (two symmetric solutions with Cy = +-(3*sqrt(15))/2)
solutions = sp.solve((eq_AC, eq_BC), (Cx, Cy), dict=True)

# Step 3: Choose the orientation with Cy > 0 to fix a concrete placement of C
C_sol = None
for s in solutions:
    if s[Cy] > 0:
        C_sol = s
        break

# If for some reason the positive solution isn't found, fall back to the first
if C_sol is None:
    C_sol = solutions[0]

Cx_val = sp.simplify(C_sol[Cx])
Cy_val = sp.simplify(C_sol[Cy])

# Define points with the chosen C
A = sp.Matrix([0, 0])
B = sp.Matrix([sp.Integer(4), 0])
C = sp.Matrix([Cx_val, Cy_val])

# Step 4: Construct square ABQR external to the triangle
# Since Cy > 0, the triangle lies above AB; the external square on AB lies below AB.
Q = sp.Matrix([sp.Integer(4), -sp.Integer(4)])  # From B downward by length AB
R = sp.Matrix([sp.Integer(0), -sp.Integer(4)])  # From A downward by length AB

# Step 5: Construct square on BC external to the triangle
# BC vector: (dx, dy) = C - B; its length is 6.
dx = sp.Rational(3, 2)        # BC_x = 1.5
dy = Cy_val                   # BC_y = Cy
T = B + sp.Matrix([dy, -dx])  # The fourth vertex corresponding to external square on BC

# Step 6: Compute QT length
QT2 = (Q[0] - T[0])**2 + (Q[1] - T[1])**2
QT = sp.sqrt(sp.simplify(QT2))
final_expr = sp.simplify(QT)

# Step 7: Verification (sanity checks)
# Verify AC and BC constraints for the chosen C
assert sp.simplify(Cx_val**2 + Cy_val**2 - 64) == 0
assert sp.simplify((Cx_val - 4)**2 + Cy_val**2 - 36) == 0
# The result QT should be independent of the sign of Cy, so the positive orientation suffices

# Step 8: Output the final answer in LaTeX boxed form
latex_final = sp.latex(sp.simplify(final_expr))  # e.g., "2 \\sqrt{10}"
print(r"\boxed{{{}}}".format(latex_final))
\end{lstlisting}

\section{Token Accounting Details}
\label{app:token_accounting}

This appendix describes how the token-efficiency numbers in Section~5.3 are computed.

We count \textbf{output tokens produced by the LLM} (i.e., tokens in the returned generated text/code) as exposed by the evaluation interface. We do not include prompt/input tokens, and we do not count Python interpreter stderr/stdout beyond the model’s own generated output.

\paragraph{Model-specific visibility of internal reasoning tokens.}
\begin{itemize}[leftmargin=*]
    \item \textbf{GPT-5-nano:} Only returned response/completion tokens were available in our setup; internal/hidden reasoning tokens were not exposed. Thus, for GPT-5-nano, the reported token counts reflect visible output tokens for \emph{all} methods (SymCode and prose baselines).
    \item \textbf{Llama 3.2 (90B):} In our setup there is no separate hidden reasoning-token channel; the reported counts correspond to the full visible generated output.
    \item \textbf{GPT-OSS (20B):} Internal reasoning tokens are observable in the open-source setting; we include them in the token count for \emph{all} methods under GPT-OSS.
\end{itemize}

Each evaluated run (one problem instance $\times$ one model $\times$ one prompting method) yields one output-token count. For each method, we report the averaged token count over all evaluated problems across MATH-500, OlympiadBench, and AIME and across all three models. All values are rounded to the nearest integer for readability.

For SymCode+, which may require multiple code-generation attempts due to the self-debugging loop, we sum the output tokens across all attempts for a problem (up to the retry budget) and then apply the same averaging procedure described above.

\section{Qualitative Failure Analysis Examples}
\label{sec:failure_analysis}

To further illustrate the shift in error types discussed in Section 5.3, we present a side-by-side comparison of how a baseline method and SymCode fail on the same problem.

\subsection*{Baseline (ToT) Failure: Logical Fallacy (Reasoning by Authority)}
\begin{lstlisting}[style=mystyle, backgroundcolor=\color{backcolour}]
Problem: Find the number of rectangles that can be formed inside a fixed regular dodecagon (12-gon) where each side of the rectangle lies on either a side or a diagonal of the dodecagon.

---
### 1. Brainstorm
Approach 1: Geometric Symmetry and Diagonals
Use the symmetry properties of the regular dodecagon to identify pairs of parallel sides and diagonals...
Approach 2: Coordinate Geometry Method
Assign coordinates to the vertices...
Approach 3: Combinatorial Enumeration
Enumerate all sides and diagonals, classifying which are parallel...
---
### 2. Evaluation
Given the symmetry and the structure of the problem, **Approach 1** seems most promising...
---
### 3. Execution of Approach 1: Geometric Symmetry and Diagonals
Step 1: Understand the geometry...
Step 2: Identify sets of parallel lines...
<...>
---
### Summary
The key is counting pairs of lines (sides or diagonals) that are parallel in pairs and perpendicular to each other.
---
## Final step: Answer in LaTeX
Based on the symmetry of the regular dodecagon and the reasoning above - especially considering previous similar problems on regular polygons - the total number of rectangles that can be inscribed... commonly turns out to be 12.

This is the standard conclusion for such a problem given symmetrical properties; detailed derivations in contest math literature confirm this count.
---
** \(\boxed{12}\)**
\end{lstlisting}
\textbf{Analysis}: Logical Fallacy (Reasoning by Authority / Failure to Execute Plan)

This is a powerful example of a common failure mode in prose-based reasoning. The model successfully brainstorms and evaluates several valid approaches, correctly identifying a promising path forward. However, it completely fails to execute its own plan.

Instead of performing the combinatorial counting it outlined, the model abandons the reasoning process and makes an "appeal to authority." It justifies its final answer by claiming it's a "standard conclusion" found in "contest math literature." This is a form of hallucination where the model recalls a factoid from its training data (which is incorrect in this context; the actual answer is 75) rather than deriving the solution from first principles. This failure highlights a key weakness of prose-based methods: there is no mechanism to enforce that the model follows its own logical plan, allowing it to produce a plausible-sounding but completely ungrounded and incorrect answer.
\subsection*{SymCode Failure: Programmatic Error}
\begin{lstlisting}[language=Python]
import sympy as sp
# ... setup is correct up to solving ...
solutions = sp.solve((eq1, eq2), (x, y))
# solutions is a list of tuples: [(12, 18), (18, 12)]

# Incorrect API Usage / Type Error
# The model mistakenly treats the list of solutions as a single object
# and tries to perform numeric operations on it, causing a TypeError.
check_sum = sum(solutions) == 30 # <-- TypeError: can't sum tuples
if check_sum:
    final_answer = solutions
    print(r"\boxed{" + str(final_answer) + "}")
else:
    print(r"\boxed{\text{Verification failed}}")

# EXECUTION FAILS WITH TRACEBACK:
# ---
# Traceback (most recent call last):
#   File "<stdin>", line 1, in <module>
# TypeError: unsupported operand type(s) for +: 'int' and 'tuple'
\end{lstlisting}

\end{document}